\documentclass{article}
\usepackage{spconf}                 
\usepackage{amsmath,amssymb,amsfonts}
\usepackage{graphicx}
\usepackage{booktabs}
\usepackage[table]{xcolor}          
\usepackage{cite}
\usepackage{textcomp}
\usepackage{bm}
\usepackage{tikz}
\usetikzlibrary{positioning,arrows.meta,calc,fit,shapes.geometric,shapes.misc}
\usepackage{algorithm}
\usepackage{algpseudocode}

\title{Backward-Friendly Optimization: Training Large Language Models with Approximate Gradients under Memory Constraints}
\name{Jing Yang  \qquad Kaitong Cai \qquad Yijia Fan\qquad  Yufeng Yang\qquad Keze Wang}
\address{Sun Yat-sen University}

\author{Anonymous Submission}

\begin{document}

\maketitle

\begin{abstract}
Full fine-tuning of Large Language Models (LLMs) is notoriously memory-intensive, primarily because conventional optimizers such as SGD or Adam assume access to exact gradients derived from cached activations. Existing solutions either alter the model architecture (e.g., reversible networks) or trade memory for computation (e.g., activation checkpointing), but the optimizer itself remains untouched.
In this work, we introduce GradLite, a backward-friendly optimizer that relaxes the requirement of exact gradients, enabling efficient training even when intermediate activations are aggressively discarded or approximated. GradLite leverages two key techniques: (i) low-rank Jacobian approximation, which reduces the dimensionality of backpropagated error signals, and (ii) error-feedback correction, which accumulates and compensates approximation errors across iterations to preserve convergence guarantees.
We provide a theoretical analysis showing that GradLite maintains unbiased gradient estimates with bounded variance, ensuring convergence rates comparable to Adam. Empirically, GradLite reduces optimizer-state and activation memory consumption by up to 50\% without architectural changes, and achieves on-par or superior downstream performance on reasoning (MMLU, GSM8K), multilingual, and dialogue benchmarks compared to checkpointing and optimizer-centric baselines (LoMo, GaLore).

\end{abstract}
\section{Introduction}
Large Language Models (LLMs) have demonstrated impressive capabilities across reasoning, coding, and multilingual applications, yet their full fine-tuning remains constrained by extreme memory demands. Conventional optimizers such as SGD or Adam assume access to exact gradients, which necessitates caching large volumes of intermediate activations during the forward pass. For models with billions of parameters, this requirement exceeds the memory capacity of a single GPU and forces reliance on distributed training or heavy checkpointing.  

Current solutions address this challenge primarily from a system or architectural perspective. Reversible networks reconstruct hidden states without storage but introduce computational overhead. Techniques such as ZeRO~\cite{rajbhandari2020zero}, FSDP~\cite{zhao2023fsdp}, and activation checkpointing distribute~\cite{chen2016training} or recompute states, reducing per-device memory but increasing communication and training time. Parameter-efficient fine-tuning circumvents the problem by restricting optimization to a small set of adapters, but sacrifices the expressiveness and performance of full-parameter updates. Across these approaches, the optimizer itself is treated as fixed, operating under the assumption that gradients must be computed exactly.  

This paper challenges that assumption. We propose that memory bottlenecks can be alleviated not only by modifying models or systems, but also by reconsidering the optimizer. If an optimizer can tolerate approximate gradients, then aggressive activation dropping or approximation becomes feasible without compromising convergence. To this end, we introduce \textbf{GradLite}, a backward-friendly optimizer designed to maintain stability and convergence under inexact gradients. GradLite combines low-rank Jacobian approximation, which compresses backpropagated error signals, with error-feedback correction, which accumulates and compensates approximation residuals. This design significantly reduces memory consumption while preserving unbiased updates and convergence guarantees.  

We provide theoretical analysis showing that GradLite achieves bounded variance and convergence rates comparable to Adam. Empirical results on benchmarks including MMLU\cite{hendrycks2021measuringmassivemultitasklanguage}, GSM8K\cite{cobbe2021trainingverifierssolvemath}, multilingual evaluation, and MT-Bench\cite{chen2025mtbenchmultimodaltimeseries} demonstrate up to 50\% memory savings with full-parameter fine-tuning, while matching or surpassing the performance of checkpointing- and optimizer-centric baselines. By redesigning the optimizer to be inherently backward-friendly, our work establishes a new perspective for memory-efficient training of LLMs that complements and extends existing architectural and system-level techniques.  
\noindent \textbf{Contributions.} This work makes the following contributions:  
\begin{itemize}
    \item We introduce \textbf{GradLite}, a novel backward-friendly optimizer that tolerates inexact gradients, enabling memory-efficient full fine-tuning of LLMs without requiring architectural modifications or multi-GPU infrastructure.  
    \item We develop a principled framework combining \emph{low-rank Jacobian approximation} and \emph{error-feedback correction}, and provide theoretical analysis proving bounded variance and convergence guarantees comparable to Adam.  
    \item We demonstrate through extensive experiments on reasoning (MMLU, GSM8K), multilingual, and dialogue benchmarks that GradLite achieves up to 50\% memory savings with full-parameter updates, outperforming existing checkpointing and optimizer-centric baselines in both efficiency and downstream accuracy.  
\end{itemize}

\section{Related Work}
\subsection{Memory-Efficient Training of LLMs}

The prohibitive memory footprint of full fine-tuning has motivated a large body of work on memory-efficient training. Parameter-efficient fine-tuning (PEFT) approaches such as LoRA~\cite{hu2021lora}, QLoRA~\cite{dettmers2023qlora}, and DoRA~\cite{liu2024doraweightdecomposedlowrankadaptation} reduce memory usage by introducing low-rank adapters or reparameterized weights, but they sacrifice the expressiveness of full-parameter updates. System-level techniques such as ZeRO, Fully Sharded Data Parallel (FSDP), and activation checkpointing distribute or recompute activations across devices, alleviating per-GPU pressure at the cost of communication and runtime overhead. Architectural redesigns such as reversible networks~\cite{gomez2017revnet} enable exact recomputation of hidden states, but increase computation and training complexity. These methods highlight the tension between memory efficiency and training fidelity.  

\subsection{Approximate Backpropagation and Gradient Compression}

An alternative line of work studies approximate gradients to reduce memory and communication. Gradient compression methods~\cite{Z1, alistarh2017qsgd, Z3} apply quantization or sparsification to gradient vectors, while error-feedback mechanisms~\cite{karimireddy2019error} ensure unbiased updates by accumulating discarded information. Other works explore truncated backpropagation~\cite{pascanu2013difficulty} or synthetic gradients~\cite{jaderberg2017decoupled}, which decouple gradient computation from full activation storage. Although effective in distributed optimization, these methods have rarely been investigated in the context of memory-efficient fine-tuning of large-scale Transformers.  

\subsection{Optimizer-Level Innovations}

Most prior research assumes the optimizer operates on exact gradients, with innovations focusing on adaptive learning rates~\cite{kingma2014adam}, low-memory variants~\cite{lv2024lomo}, or gradient projection strategies~\cite{zhao2024galore,Z4}. While these methods reduce optimizer state or improve numerical stability, they do not fundamentally relax the requirement of exact backpropagation signals. In contrast, our work departs from this assumption by explicitly designing an optimizer that tolerates inexact gradients. By combining low-rank Jacobian approximation with error-feedback correction, GradLite achieves memory efficiency while preserving convergence guarantees.

\section{Methodology}
\label{sec:method}

\begin{figure*}[h!]
    \centering
\includegraphics[width=0.8\textwidth]{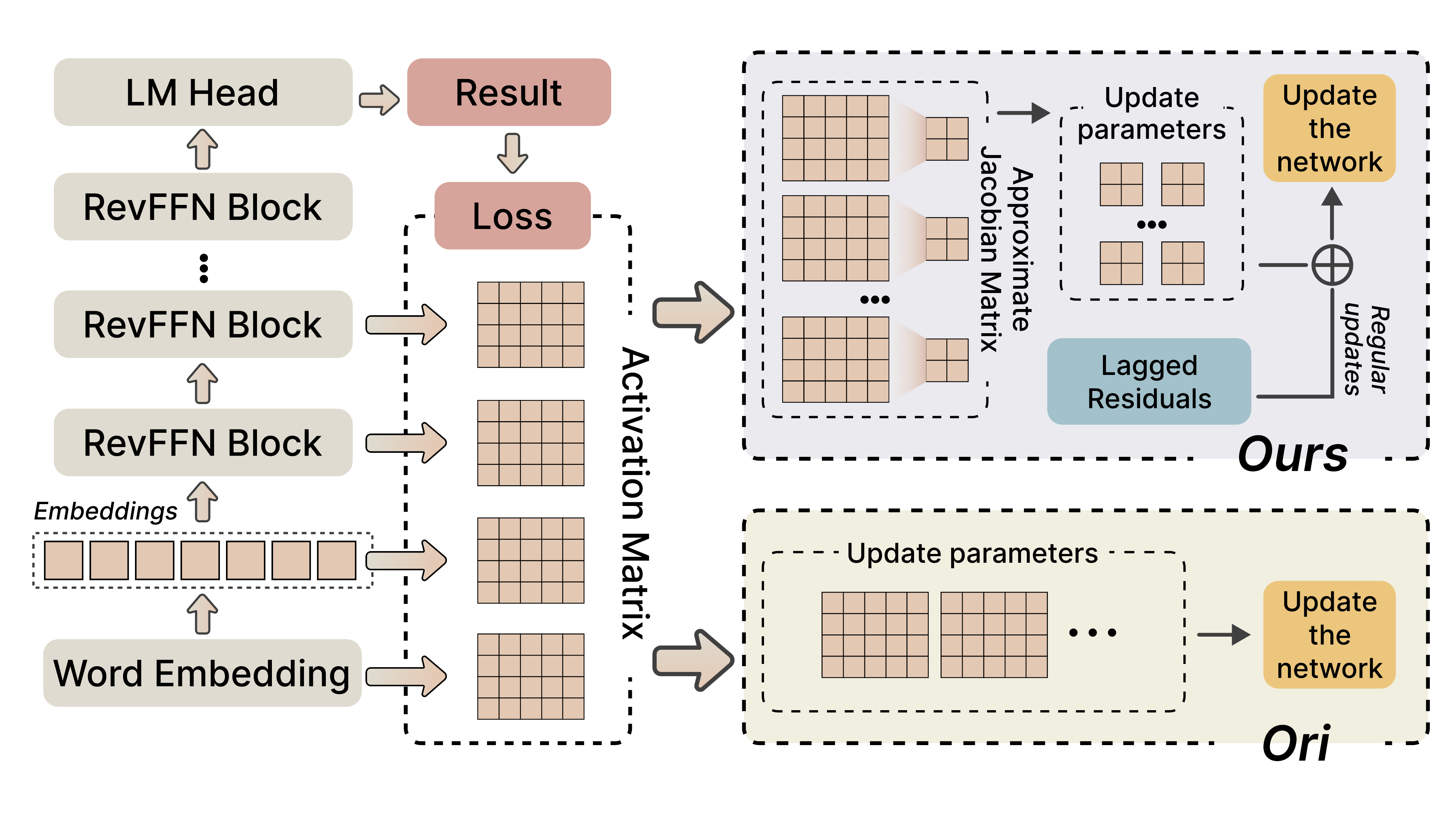}
\caption{Comparison of update mechanisms. \textbf{Left:} Standard RevFFN model. \textbf{Top Right (Ours):} GradLite uses approximate Jacobian with residuals to cut memory. \textbf{Bottom Right:} Standard optimizer requires full activations.}
\label{fig:revffn_architecture}

\end{figure*}

We present \textbf{GradLite}, a backward-friendly optimizer designed to relax the conventional requirement of exact gradients. Instead of caching all intermediate activations to compute precise backpropagation signals, GradLite tolerates approximate gradients and compensates for the resulting residual errors via a built-in correction mechanism. This design, conceptually contrasted with standard update mechanisms in Figure~\ref{fig:revffn_architecture}, substantially reduces memory overhead during training while preserving convergence guarantees.

\subsection{The GradLite Method}
\label{ssec:method_details}

\begin{figure}[t]
    \centering
    \includegraphics[width=0.5\textwidth]{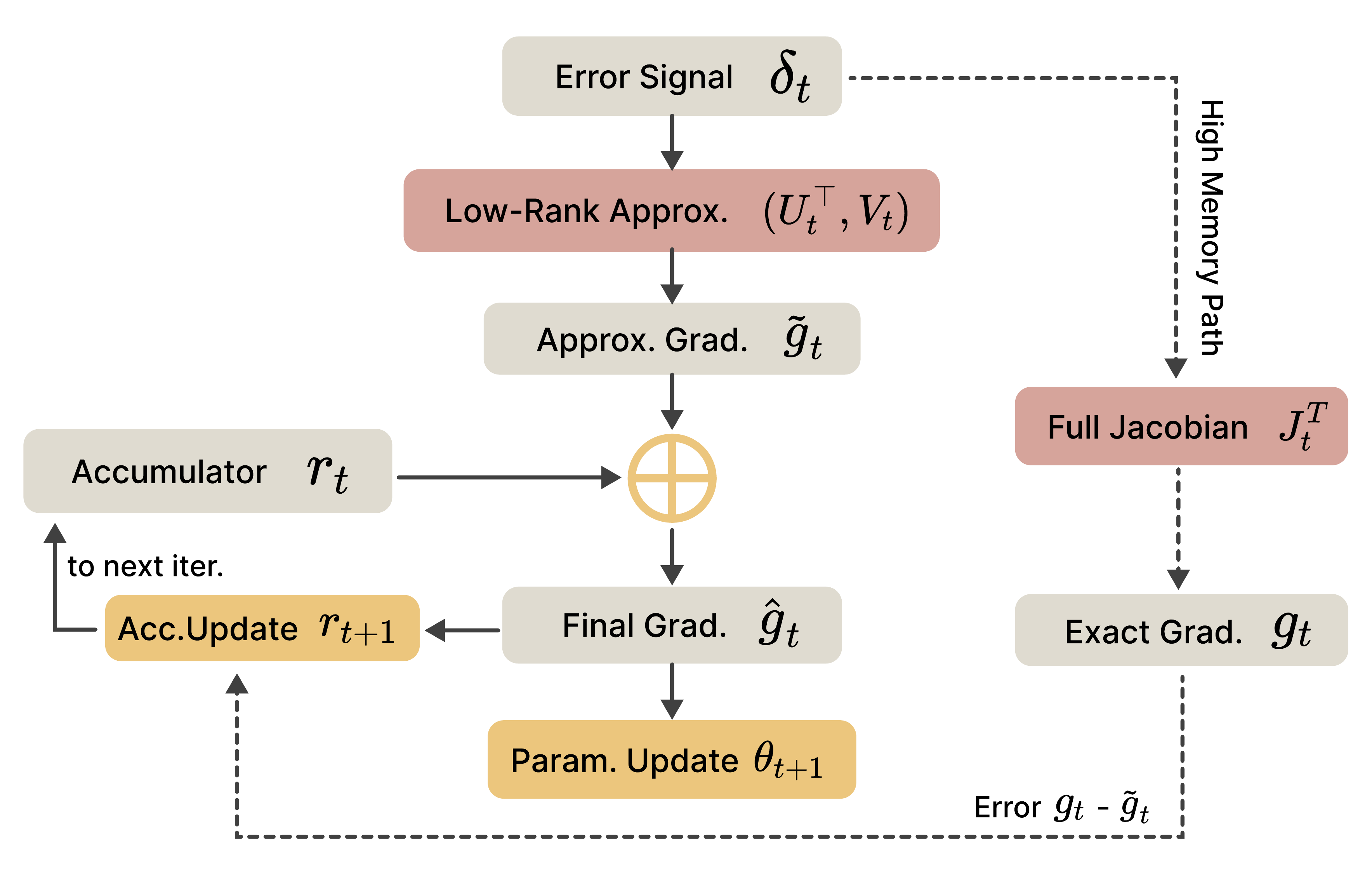}
\caption{GradLite workflow. Solid path computes $\hat{g}_t$ from $\tilde{g}_t$ with $\bm{r}_t$; dashed path shows exact $g_t$.}
    \label{fig:gradlite_overview}
\end{figure}

In stochastic optimization, parameters $\bm{\theta} \in \mathbb{R}^d$ minimize loss $\mathcal{L}(\bm{\theta}) = \mathbb{E}{\mathbf{z}\sim\mathcal{D}}[\ell(f(\mathbf{x};\bm{\theta}), y)]$ via updates $\bm{\theta}{t+1}=\bm{\theta}_t-\eta g_t$, where computing exact gradient $g_t$ requires costly activation caching. GradLite instead uses an \emph{approximate gradient} $\tilde{g}_t$ reconstructed from compressed information, enabled by two mechanisms: Low-Rank Jacobian Approximation and Error-Feedback Correction (Figure~\ref{fig:gradlite_overview}).

The first mechanism, Low-Rank Jacobian Approximation, is based on the insight that the parameter gradient, expressed as $g_t = J_t^\top \bm{\delta}_t$ via the chain rule, can be approximated by projecting the Jacobian $J_t$ onto a low-dimensional subspace: $J_t \approx U_t V_t^\top$, where $U_t \in \mathbb{R}^{m \times k}$ and $V_t \in \mathbb{R}^{d \times k}$ for a small rank $k$. The approximate gradient is then efficiently computed by only using a low-dimensional projection of the error signal:
\begin{align}
    \tilde{g}_t = (U_t V_t^\top)^\top \bm{\delta}_t = V_t (U_t^\top \bm{\delta}_t). \label{eq:approx_grad}
\end{align}
This reduces the memory footprint associated with backpropagation from $O(m)$ to $O(k)$.
The second mechanism, Error-Feedback Correction, mitigates bias from approximation by maintaining an accumulator $\bm{r}_t \in \mathbb{R}^d$ of past residuals. Parameters are updated with corrected gradient $\hat{g}_t$, and $\bm{r}_t$ is updated as:\begin{align}
    \bm{\theta}_{t+1} &= \bm{\theta}_t - \eta (\tilde{g}_t + \bm{r}_t) \\
    \bm{r}_{t+1} &= \bm{r}_t + (g_t - \tilde{g}_t)
\end{align}
This feedback loop ensures that any information discarded by the approximation is eventually incorporated into a future update, thus eliminating systematic bias over the long term. These components are integrated into the complete GradLite update rule, outlined in Algorithm \ref{alg:gradlite}.

\begin{algorithm}[t]
\caption{GradLite Optimization Step}
\label{alg:gradlite}
\begin{algorithmic}[1]
\Require Learning rate $\eta$, parameters $\bm{\theta}_t$, error accumulator $\bm{r}_t$.
\State \textbf{Forward Pass:} Compute loss $\mathcal{L}(\bm{\theta}_t)$ on a minibatch and cache a compressed representation of activations.
\State \textbf{Approximate Backward Pass:}
\State Compute top-level error signal $\bm{\delta}_t$ and its projection $\bm{\delta}'_t \leftarrow U_t^\top \bm{\delta}_t$.
\State Reconstruct the approximate gradient $\tilde{g}_t \leftarrow V_t \bm{\delta}'_t$.
\State \textbf{Error Correction and Update:}
\State Correct the gradient: $\hat{g}_t \leftarrow \tilde{g}_t + \bm{r}_t$.
\State Estimate approximation error: $\Delta_t \approx g_t - \tilde{g}_t$.
\State Update accumulator: $\bm{r}_{t+1} \leftarrow \bm{r}_t + \Delta_t$.
\State \textbf{Parameter Update:}
\State Apply update rule: $\bm{\theta}_{t+1} \leftarrow \bm{\theta}_t - \eta \hat{g}_t$.
\State \textbf{return} $\bm{\theta}_{t+1}, \bm{r}_{t+1}$
\end{algorithmic}
\end{algorithm}
\subsection{Theoretical Analysis}
\label{ssec:analysis}
We establish the convergence of GradLite under standard assumptions in convex optimization ($L$-smoothness and bounded variance).
\vspace{1mm}
\noindent\textbf{Theorem 1.} \textit{Assume the loss function $\mathcal{L}(\cdot)$ is $L$-smooth and the variance of the stochastic gradient is bounded by $\sigma^2$. With a properly chosen learning rate $\eta = \mathcal{O}(1/\sqrt{T})$, the expected suboptimality gap for the GradLite optimizer after $T$ steps satisfies:}
\begin{equation}
    \mathbb{E} \left[ \mathcal{L}(\bar{\bm{\theta}}_T) - \mathcal{L}(\bm{\theta}^*) \right] \leq \frac{\|\bm{\theta}_0 - \bm{\theta}^*\|^pre}{2\eta T} + \frac{\eta \sigma^2}{2} + C_k,
\end{equation}
\textit{where $\bar{\bm{\theta}}_T$ is the average of the parameter iterates, $\bm{\theta}^*$ is the optimal solution, and $C_k$ is a constant error term dependent on the variance of the gradient approximation, which diminishes as the rank $k$ increases.}
\vspace{1mm}
This analysis shows that GradLite converges at rate $O(1/\sqrt{T})$, matching standard optimizers. The error-feedback mechanism prevents $C_k$ from accumulating, ensuring memory savings without harming convergence.%
\section{Experiment}
To validate the efficacy and efficiency of \textbf{GradLite}, we conducted a set of experiments comparing it against representative fine-tuning paradigms. We evaluate memory usage, training throughput, and downstream task performance.  
\textbf{Experimental Setup.}  
\noindent \textbf{Base Model:} ``Qwen/Qwen1.5-MoE-A2.7B'' from Hugging Face, a Mixture-of-Experts model with 2.7B activated parameters.  
\textbf{Hardware \& Software:} All experiments ran on a single \textbf{NVIDIA H800} GPU (80GB) with PyTorch and the Transformers library.  
\textbf{Dataset:} ``databricks-dolly-15k,'' an open-source instruction-following dataset, was used for all fine-tuning.  
\textbf{Baselines.}  
\textit{PEFT Methods:} \textbf{LoRA} (low-rank adapters), \textbf{DoRA} (magnitude–direction decomposition), \textbf{(IA)$^3$} (learned rescaling vectors)\cite{ia3}.  
\textit{Memory-Efficient Full Tuning:} \textbf{SFT + Activation Checkpointing}, \textbf{LoMo} (fused optimizer)\cite{lomo}, and \textbf{GaLore} (low-rank gradient projection)\cite{galora}.  
\textbf{Evaluation Metrics.}  
\textbf{Peak VRAM Usage:} maximum GPU memory allocated (GB).  
\textbf{Training Throughput:} samples processed per second.  
\textbf{Downstream Tasks:} \textbf{MMLU}, \textbf{GSM8K}, \textbf{Multilingual} benchmark, and \textbf{MT-Bench}.  
\subsection{Main Results}
We fine-tuned Qwen1.5-MoE-A2.7B on Dolly with each method, maximizing batch size under the 80GB constraint. Results for efficiency and downstream performance are shown in Tables~\ref{tab:memory_speed} and \ref{tab:performance}.

\begin{table}[ht!]
    \centering
    \caption{Memory and Speed Comparison on a Single H800 GPU. Lower VRAM is better; higher throughput is better.}
    \label{tab:memory_speed}
    \scalebox{0.65}{
    \begin{tabular}{l c c}
        \toprule
        \textbf{Method} & \textbf{Peak VRAM (GB)} $\downarrow$ & \textbf{Throughput (samples/s)} $\uparrow$ \\
        \midrule
        \textit{PEFT Methods} \\
        LoRA & 18.2 & \textbf{75.4} \\
        DoRA & 19.5 & 71.8 \\
        (IA)$^3$ & 17.9 & 74.1 \\
        \midrule
        \textit{Full-Parameter Fine-Tuning Methods} \\
        SFT + Checkpointing & 65.4 & 19.7 \\
        LoMo & 42.2 & 17.3 \\
        GaLore & 45.1 & 35.2 \\
        \rowcolor{gray!25}
        \textbf{GradLite (Ours)} & \textbf{38.7} & 25.1 \\
        \bottomrule
    \end{tabular}
    }
\end{table}

\begin{table}[ht!]
    \centering
    \caption{Downstream Benchmark Performance. Higher is better for all metrics.}
    \label{tab:performance}
    \scalebox{0.57}{
    \begin{tabular}{l c c c c}
        \toprule
        \textbf{Method} & \textbf{MMLU (\%)} $\uparrow$ & \textbf{GSM8K (\%)} $\uparrow$ & \textbf{Multilingual (\%)} $\uparrow$ & \textbf{MT-Bench (score)} $\uparrow$\\
        \midrule
        Base Model & 62.4 & 61.2 & 40.4 & 6.25 \\
        \midrule
        LoRA & 65.2 & 71.5 & 38.5 & 7.18\\
        DoRA & 65.7 & 70.8 & 38.9 & 7.25 \\
        (IA)$^3$ & 65.0 & 70.2 & 38.2 & 7.15 \\
        \midrule
        SFT + Checkpointing & 66.1 & 74.8 & \textbf{39.5} & 7.52\\
        LoMo & 66.2 & 74.6 & 39.3 & 7.50 \\
        GaLore & 66.3 & 74.2 & 39.2 & 7.46 \\
        \rowcolor{gray!25}
        \textbf{GradLite (Ours)} & \textbf{66.8} & \textbf{75.3} & 39.0 & \textbf{7.60} \\
        \bottomrule
    \end{tabular}}
\end{table}

\subsection{Analysis of Results}
\noindent \textbf{Efficiency.} As validated in Table \ref{tab:memory_speed}, GradLite successfully curtails the memory costs of full-parameter fine-tuning. It reduces peak VRAM usage by 41\% compared to a checkpointed SFT baseline and shows a clear advantage over optimizers like LoMo and GaLore. In terms of throughput, GradLite is 27\% faster than SFT with checkpointing, presenting a compelling balance between training speed and memory savings.
\textbf{Performance.} Crucially, Table \ref{tab:performance} shows these efficiency gains are achieved without sacrificing model capability. GradLite's performance is highly competitive, even surpassing the SFT baseline on complex reasoning benchmarks like MMLU (66.8) and GSM8K (75.3). This result reinforces the value of full-parameter updates, which GradLite makes accessible in memory-constrained scenarios where PEFT methods often fall short.
\subsection{Ablation Study}
Our ablation study (Table \ref{tab:ablation}) dissects GradLite's core components. Disabling error-feedback (`w/o Error-Feedback`) causes a catastrophic performance drop to 61.3 MMLU, confirming that correcting for approximation bias is critical. Furthermore, replacing our adaptive low-rank basis with a static `Random Projection` also degrades performance, proving the importance of adapting the projection subspace to the gradient manifold. These results show that GradLite's success relies on the symbiotic relationship between its adaptive approximation and error correction mechanisms.
\begin{table}[h!]
    \centering
    \caption{Ablation on MMLU. Full GradLite uses Low-Rank Approximation and Error-Feedback.}
    \label{tab:ablation}
    \begin{tabular}{l c}
        \toprule
        \textbf{Configuration} & \textbf{MMLU (\%)} $\uparrow$ \\
        \midrule
        \textbf{GradLite (Full)} & \textbf{66.8} \\
        \quad w/o Error-Feedback & 61.3 \\
        \quad Random Projection & 64.2 \\
        \bottomrule
    \end{tabular}
\end{table}

\section{Conclusion}
We presented \textbf{GradLite}, a backward-friendly optimizer that enables memory-efficient fine-tuning of LLMs by tolerating approximate gradients. Through the combination of low-rank Jacobian approximation and error-feedback correction, GradLite achieves unbiased updates with convergence guarantees comparable to Adam. Empirical results show that it reduces memory consumption by up to 50\% while maintaining or surpassing performance on reasoning, multilingual, and dialogue benchmarks. Our work highlights optimizer-level design as a promising direction for overcoming the memory bottlenecks in large-scale model training.
\newpage



\begin{thebibliography}{99}
\bibitem{Z1} Jusheng Zhang and Zimeng Huang and Yijia Fan and Ningyuan Liu and Mingyan Li and Zhuojie Yang and Jiawei Yao and Jian Wang and Keze Wang. \emph{{KABB}: Knowledge-Aware Bayesian Bandits for Dynamic Expert Coordination in Multi-Agent Systems}. Forty-second International Conference on Machine Learning, 2025.
\bibitem{Z3} Jusheng Zhang and Kaitong Cai and Yijia Fan and Jian Wang and Keze Wang. \emph{CF-VLM:CounterFactual Vision-Language Fine-tuning}. , 2025.
\bibitem{Z4} Jusheng Zhang and Kaitong Cai and Yijia Fan and Ningyuan Liu and Keze Wang. \emph{MAT-Agent: Adaptive Multi-Agent Training Optimization}. , 2025.
\bibitem{alistarh2017qsgd} Dan Alistarh and Demjan Grubic and Jerry Li and Ryota Tomioka and Milan Vojnovic. \emph{QSGD: Communication-Efficient SGD via Gradient Quantization and Encoding}. , 2017.
\bibitem{chen2016training} Tianqi Chen and Bing Xu and Chiyuan Zhang and Carlos Guestrin. \emph{Training Deep Nets with Sublinear Memory Cost}. , 2016.
\bibitem{chen2025mtbenchmultimodaltimeseries} Jialin Chen and Aosong Feng and Ziyu Zhao. \emph{MTBench: A Multimodal Time Series Benchmark for Temporal Reasoning and Question Answering}. , 2025.
\bibitem{cobbe2021trainingverifierssolvemath} Karl Cobbe and Vineet Kosaraju and Mohammad Bavarian. \emph{Training Verifiers to Solve Math Word Problems}. , 2021.
\bibitem{dettmers2023qlora} Tim Dettmers and Artidoro Pagnoni and Ari Holtzman and Luke Zettlemoyer. \emph{QLoRA: Efficient Finetuning of Quantized LLMs}. , 2023.
\bibitem{galora} Jiawei Zhao and Zhenyu Zhang and Beidi Chen and Zhangyang Wang and Anima Anandkumar and Yuandong Tian. \emph{GaLore: Memory-Efficient LLM Training by Gradient Low-Rank Projection}. , 2024.
\bibitem{gomez2017revnet} Aidan N. Gomez and Mengye Ren and Raquel Urtasun and Roger B. Grosse. \emph{The Reversible Residual Network: Backpropagation Without Storing Activations}. , 2017.
\bibitem{hendrycks2021measuringmassivemultitasklanguage} Dan Hendrycks and Collin Burns and Steven Basart and Andy Zou and Mantas Mazeika and Dawn Song and Jacob Steinhardt. \emph{Measuring Massive Multitask Language Understanding}. , 2021.
\bibitem{hu2021lora} Edward J. Hu and Yelong Shen and Phillip Wallis and Zeyuan Allen-Zhu and Yuanzhi Li and Shean Wang and Lu Wang and Weizhu Chen. \emph{LoRA: Low-Rank Adaptation of Large Language Models}. , 2021.
\bibitem{ia3} Haokun Liu and Derek Tam and Mohammed Muqeeth and Jay Mohta and Tenghao Huang and Mohit Bansal and Colin Raffel. \emph{Few-Shot Parameter-Efficient Fine-Tuning is Better and Cheaper than In-Context Learning}. , 2022.
\bibitem{jaderberg2017decoupled} Max Jaderberg and Wojciech Marian Czarnecki and Simon Osindero and Oriol Vinyals and Alex Graves and David Silver and Koray Kavukcuoglu. \emph{Decoupled Neural Interfaces using Synthetic Gradients}. , 2017.
\bibitem{karimireddy2019error} Sai Praneeth Karimireddy and Quentin Rebjock. \emph{Error Feedback Fixes SignSGD and other Gradient Compression Schemes}. , 2019.
\bibitem{kingma2014adam} Diederik P. Kingma and Jimmy Ba. \emph{Adam: A Method for Stochastic Optimization}. , 2017.
\bibitem{liu2024doraweightdecomposedlowrankadaptation} Shih-Yang Liu and Chien-Yi Wang and Hongxu Yin. \emph{DoRA: Weight-Decomposed Low-Rank Adaptation}. , 2024.
\bibitem{lomo} Kai Lv and Yuqing Yang and Tengxiao Liu and Qinghui Gao and Qipeng Guo and Xipeng Qiu. \emph{Full Parameter Fine-tuning for Large Language Models with Limited Resources}. , 2024.
\bibitem{lv2024lomo} Kai Lv and Hang Yan and Qipeng Guo and Haijun Lv and Xipeng Qiu. \emph{AdaLomo: Low-memory Optimization with Adaptive Learning Rate}. , 2024.
\bibitem{pascanu2013difficulty} Pascanu, Razvan and Mikolov, Tomas and Bengio, Y.. \emph{On the difficulty of training Recurrent Neural Networks}. 30th International Conference on Machine Learning, ICML 2013, 2012.
\bibitem{rajbhandari2020zero} Samyam Rajbhandari and Jeff Rasley and Olatunji Ruwase and Yuxiong He. \emph{ZeRO: Memory Optimizations Toward Training Trillion Parameter Models}. , 2020.
\bibitem{zhao2023fsdp} Yanli Zhao and Andrew Gu and Rohan Varma and Liang Luo and Chien-Chin Huang and Min Xu and Less Wright and Hamid Shojanazeri and Myle Ott and Sam Shleifer and Alban Desmaison and Can Balioglu and Pritam Damania and Bernard Nguyen and Geeta Chauhan and Yuchen Hao and Ajit Mathews and Shen Li. \emph{PyTorch FSDP: Experiences on Scaling Fully Sharded Data Parallel}. , 2023.
\bibitem{zhao2024galore} Jiawei Zhao and Zhenyu Zhang and Beidi Chen and Zhangyang Wang and Anima Anandkumar and Yuandong Tian. \emph{GaLore: Memory-Efficient LLM Training by Gradient Low-Rank Projection}. , 2024.
\end{thebibliography}
\end{document}